\documentclass{cta-author}

{}
{}
{}

\usepackage{hyperref}
\hypersetup{
    colorlinks=true,
    linkcolor=blue,
    filecolor=magenta,      
    urlcolor=cyan,
}
 
\begin{document}

\supertitle{IET Research Journal Paper}

\title{
A Unified Deep Learning Approach for Prediction of Parkinson's Disease}
\author{\au{James Wingate$^{1}$}, \au{Ilianna Kollia$^{2}$}, \au{Luc Bidaut $^{1}$}, \au{Stefanos Kollias$^{1,2}$}}

\address{\add{1}{School of Computer Science, University of Lincoln, Brayford Pool, LN6 7TS, Lincoln, UK}
\add{2}{School of Electrical and Computer Engineering, National Technical University of Athens, 9 Iroon Polytechniou street, Zografou, 15780, Athens, Greece}
\email{JWingate@lincoln.ac.uk, ilianna2@mail.ntua.gr}}

\begin{abstract}
The paper presents a novel approach, based on deep learning, for diagnosis of Parkinson's disease through medical imaging. The approach includes analysis and use of the knowledge extracted by Deep Convolutional and Recurrent Neural Networks (DNNs) when trained with medical images, such as  Magnetic Resonance Images and DaTscans. Internal representations of the trained DNNs constitute the extracted knowledge which is used in a transfer learning and domain adaptation manner, so as to create a unified framework for prediction of Parkinson's across different medical environments. A large experimental study is presented illustrating the ability of the proposed approach to effectively predict Parkinson's, using different medical image sets from real environments.
\end{abstract}

\maketitle

\section{Introduction}\label{sec1}
Current biomedical signal analysis, including medical imaging, has been for long based on feature extraction combined with quantitative and qualitative processing. Recent advances in Machine Learning (ML) and Deep Neural Networks (DNNs) have provided state-of-the-art performance in major signal processing tasks, such as computer vision, speech recognition, human computer interaction and natural language processing. DNNs can be trained as end-to-end-architectures which include different network types and provide numerical or symbolic outputs \cite{goodfellow16}. Medical diagnosis is an area in which ML and DNNs can be effectively used. This is due to their ability to analyse big amounts of data, signals, images and image sequences, to find patterns in them and to use them for effective classification, regression and prediction purposes. Various promising results have been obtained in a variety of problems \cite{sadja06, azizi17, li14}.

Parkinson's is one of the most common neurodegenerative disorders among people from 50 to 70 years old, especially in countries with elderly population, such as United States and the European Union. Early prediction is crucial for assisting patients to retain a good quality of life. Therefore, developing techniques that are able to provide accurate and trustworthy predictions of Parkinson's in subjects is of major significance for generating a society that cares about people's well being.

Prediction of Parkinson's \cite{goetz08, hoehn11} can be based on analysis of medical images, in particular Magnetic Resonance Images (MRIs) and Dopamine Transporters scans (DaTscans).  MRI analysis targets detection of variations in brain areas, especially examining the volume of the surface of substantia nigra, the lenticular nucleus and the head of the caudate nucleus. DaTscans are produced by single photon emission computer tomography (SPECT), with 123-I-Ioflupane being provided to the patients. DaTscans are used for detecting whether there is degeneration of dopaminergic neurons in the substantia nigra. For diagnosis of Parkinson's, doctors focus on the images and scans that are most representative, select the areas around caudate nucleus head, make comparison with the cerebellum, calculate and use ratios of defined volumes for making their prediction.

Machine learning and classification methods \cite{das10} have been used  for diagnosis of Parkinson's based on MRIs \cite{salvatore14}, or DaTscans \cite{rojas13} in the last decade. Recent developments in deep learning have provided further progress in this direction. Deep Convolutional and Recurrent Neural Networks (CNNs, CNN-RNNs) have been developed and used for prediction of Parkinson's \cite{kollias18} achieving high prediction accuracy, based on a new Parkinson's database including MRI and DaTscan image data \cite{tagaris18}. 

However, although deep neural networks are capable of analysing complex data, they  lack transparency in their decision making, in the sense that it is not straightforward to justify their prediction, or to visualize the features on which the decision was based. Moreover, they generally require large amounts of data in order to learn and become able to adapt to different medical environments, or different patient cases. This makes their use difficult in healthcare, where trust and personalisation are key issues. 

In this paper we adopt the DNN architecture developed in \cite{kollias18} as a model that can potentially be applied to other medical environments, or respective datasets. However, the latter generally include medical images with different characteristics, e.g., scans can be color or gray-scale, they may have different size, or there can be different numbers of images per subject. As a consequence, direct application of the trained DNN to other datasets is not generally successful. 

Various methods can be used to face this problem. Training the DNN model from scratch with each new dataset is a possibility, but the result would be to create many different DNNs solving the same problem, but for different data cases. No interoperability would be feasible among them. Merging all possible datasets, so that a single DNN is trained on all of them would be another possibility, but this is rather unfeasible, due to both implementation and privacy issues. 

Transfer learning is another approach, usually adopted in deep learning methodologies \cite{tan18, kollias18-2}, according to which the DNN model trained with the original dataset is used to initialize DNN re-training with the new dataset. However, a serious problem arises: as the refined DNN learns to predict over the new dataset, it tends to forget the old data that are not used in the retraining procedure. As in learning from scratch, local than global prediction models would be generated.

In the following, we propose a novel approach that is able to overcome the above mentioned shortcomings and problems, providing a unified prediction model for Parkinson's based on DaTscans and/or MRI data. 

At first we extract appropriate internal features, say features $\textbf v$, from the DNN model trained with the dataset developed in \cite{tagaris18}. Using a clustering methodology, we generate concise 
representations, say $\textbf c$, of these features, which are annotated by medical experts to denote patient or non-patient categories. These representations are used in the proposed approach, in an efficient and transparent way, based on the nearest neighbour criterion, to predict whether a new subject's data indicate a Parkinson's status, or not.   
 
A novel approach is then presented, for training a DNN model with new subjects' data, in particular with different datasets, alleviating the catastrophic forgetting problem which was mentioned above. 

This approach includes a transfer learning step, in which
 we apply the  originally trained DNN to all data of the new dataset, deriving a corresponding set of features. It is this set of features, which has been extracted using the knowledge of the former DNN, that we use as training data for a new DNN model so as to learn to predict Parkinson's on the new dataset. Then, based on the new trained DNN model we extract a new set of features, say $\textbf v'$ and a concise representation $\textbf c'$. The unified Parkinson's prediction model is produced by merging the $\textbf c$ and $\textbf c'$ representation sets. 
 Having achieved high precision and recall metrics in the derivation of each one of these representations ensures that the generated unified model provides high prediction accuracy in the derived representation space.    
 
Another issue is then examined in this paper, showing that the proposed approach can be used to improve Parkinson's prediction in cases and environments where some input data types, e.g., DaTscans, are not available and prediction is made only through MRI analysis. A domain adaptation methodology for new DNN training is presented, which uses a novel error criterion based on the above-described $\textbf c$ representations. 

An extensive experimental study is presented in the paper, which develops, adapts and evaluates DNNs in all above scenarios, illustrating the excellent performance achieved in them. Two different databases are used for this purpose: the database described in \cite{tagaris18} and the Parkinson's Progression Markers Initiative (PPMI) database \cite{marek11}, which include MRI and DaTscan data, as well as textual information from patients and controls.  

Section II provides a presentation of the above two databases which we use in this paper; it also presents related work mainly focusing on methods that have been recently applied to these databases. 
Section III describes
the derivation of the above-mentioned $\textbf v$ and $\textbf c$ representations from a DNN architecture trained on a given dataset, such as the one in \cite{tagaris18}. 
Section IV presents the use of these representations for training a new DNN model with a new dataset, such as the PPMI dataset, finally deriving the unified prediction model.

Section V describes the proposed  domain adaptation methodology for using the obtained knowledge to improve Parkinson's prediction in environments with less facilities, e.g., lacking DaTscan equipment and respective information.
 Section VI presents the experimental study, evaluating the proposed approaches in the above mentioned real datasets. Section VII provides the conclusions and the directions of our future work. 

\section{Related Work}\label{sec2}
The Parkinson's Progression Markers Initiative (PPMI) dataset has been created through collaboration of researchers, funders, and study participants so as improve Parkinson's Disease therapeutics via the identification of progression biomarkers. The PPMI study includes a cohort of: 423 patients with Parkinson's disease (PD), who have been diagnosed for two years or less and do not take PD medications; 196 control subjects; 64 subjects who have been consented as PD, but whose DaTscans do not reveal dopaminergic deficit (SWEDD). Other subject categories, such as prodromal ones, or subjects with genetic mutations are also followed in the study. There is at least one Datscan, in the form of gray scale image, as well as MRI for each subject.

A variety of techniques have been applied to the PPMI dataset. During the last three years, machine learning techniques, such as Support Vector Machines (SVMs), logistic regression, random forests (RFs), and decision trees have been used for PD diagnosis. Such methods have been applied to patient questionnaires \cite{prashanth18}, reporting an accuracy over 95 \%. They were also used to analyse extracted features (related to uptake ratios on the striatum, volume and length of the striatal area) from 652 DaTscans \cite{oliveira18}, reporting an accuracy of 97.9 \%, or other features from the Unified PD Rating Scale \cite{prashanth18-2}, reporting an accuracy of 97.46 \%. 

Machine learning techniques, i.e., SVMs and RFs, were also applied to features extracted from MRI data \cite{amoroso18}, reporting an accuracy ranging from 88 \% to 93 \%, in which clinical features were also considered, apart from network features. 

Techniques based on Self-Organizing Maps (SOMs) combined with SVMs have been used to understand the pathology and provide PD diagnosis \cite{singh18}, reporting an accuracy of about 95.4 \%. Techniques using Fisher's linear discriminant analysis and locality preserving projection for feature selection, as well as a multitask framework, have been applied to discriminate among PD, control and SWEDD subjects \cite{lei18, lei18-2}, reporting accuracy about 84 \%.  

Use of Tensorflow as an interface for PD diagnosis based on medical imaging has been proposed \cite{zhang17}, using a neural network model and providing an accuracy of 97.34 \%.

Another Parkinson's database has been recently developed \cite{tagaris18}, based on anonymised data from 75 subjects, 50 subjects with PD and 25 controls with Parkinson-related syndromes, of the Georgios Gennimatas Hospital in Athens, Greece. It includes at least one DaTscan, in the form of colour image, and many MRI per subject. In total, it includes 925 DaTscans, 595 of which come from subjects with PD and 330 from controls; and 41528 MRIs, 31147 of which represent PD and 10381 non-PD. 

Deep neural networks, including Convolutionsal (CNNs), Convolutional and Recurrent (CNN-RNNs) have been developed in \cite{kollias18, tagaris17} for PD prediction using the DaTscan and MRI data included in the above-mentioned database \cite{tagaris18}. 

\begin{figure}
    \begin{center}
        \includegraphics[scale=0.35]{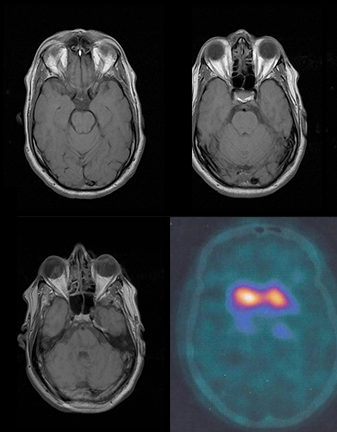}
        \caption{A DNN input including a DaTscan and three consecutive Magnetic Resonance Images from the dataset \cite{tagaris18}}
        \label{fig:ntua1}
    \end{center}
\end{figure}

In contrast to most of the techniques which were applied to the PPMI dataset, DNNs do not require a feature selection step, since features are automatically detected and extracted during DNN training. The DaTscans and/or the MRI images were directly presented at the input of the DNN.  In order to extract volumetric information, the MRI input data were provided to the DNN in consecutive triplets.  As a consequence, the DNN inputs consisted of a colour  DaTscan and/or three consecutive MRIs. An input sample, including a DaTscan and an MRI triplet is shown in Fig. \ref{fig:ntua1}. Moreover, to tackle imbalanced data between the two categories, a data augmentation strategy has been used \cite{goodfellow16, kollias18}, rising the number of combined, i.e., DaTscan and MRI inputs to a balanced number of 150.000 inputs.  

DNN training was implemented by using the pre-trained ResNet-50 structure \cite{he16}, transfer learning and adaptation \cite{kollias17, ng15} of its convolutional layers' weights, followed by training the fully connected layers and the recurrent part of the architecture; the latter was composed of gated recurrent units \cite{chung14}.  

Experiments have been presented \cite{kollias18, tagaris18} comparing the obtained accuracy, when feeding the DNNs with only DaTscan inputs, or with only MRI inputs, or with both DaTscans and MRI inputs. By training CNN and CNN-RNN architectures with the resulting dataset, a highest accuracy of 98 \% was achieved when using both types of data as inputs. An accuracy of 94 \% was achieved when using only DaTscan inputs, while a much lower accuracy of 70 \% was obtained when using only MRI inputs.

In the following, we extend this DNN architecture, as well as some early results on extraction of latent information from it which we recently presented in \cite{kollia19}, to derive a unified prediction model, which can be effectively and efficiently applied for PD diagnosis across both the database \cite{tagaris18} and the PPMI dataset, overcoming the DNN shortcomings described in the previous Section. 

\begin{figure}
    \begin{center}
        \includegraphics[scale=0.35]{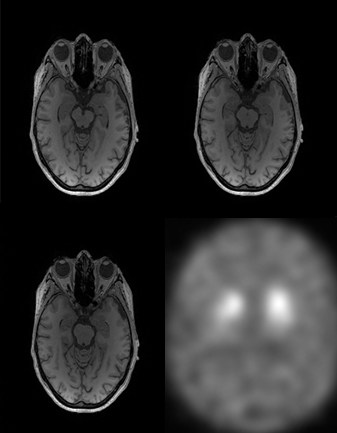}
        \caption{A DNN input including a DaTscan and three consecutive Magnetic Resonance Images from the PPMI dataset \cite{marek11}}
        \label{fig:ppmi1}
    \end{center}
\end{figure}

For comparison purposes, Fig. \ref{fig:ppmi1} shows a respective input from the PPMI dataset, including a DaTscan and a triplet of MRIs. It can be seen that the daTscans in this database are gray scale images, in contrast to the colour scans of the former database. 

\section{Extracting Concise Representations from Trained Deep Neural Networks}\label{sec3}

Our approach starts by training a deep neural architecture, such as a convolutional, or convolutional-recurrent network one, to analyse medical images, i.e., DaTscans and/or MRI images, collected in a specific medical centre, or hospital (as in \cite{tagaris18}), for predicting the status (PD, or non-PD) of their subjects. 

 As in \cite{kollias18} we consider a CNN part that can have, either a well-known structure, such as ResNet-50, on be generally composed of convolutional and pooling layers, followed by a small number of fully-connected layers. ReLU neuron models are used in this part. In the case of a convolutional and recurrent network, a small number of hidden layers with Long Short Term Memory (LSTM) neuron models, or Gated Recurrent Units (GRU) are used on top of the CNN part, providing the final classification, or prediction, outputs.
 
Research has recently focused on extracting trained DNN representations and use them for classification purposes \cite{azizi17, kollias17}, either by an auto-encoder methodology, or by monitoring neuron outputs in the convolutional, or/and fully connected network layers. 

In our approach we select to extract and further analyse the, say $M$ outputs of the last fully connected  layer, or last hidden layer of the trained CNN, or CNN-RNN respectively.  This is due to the fact that these outputs constitute high level, semantic extracts, based on which the trained DNN provides its final predictions. Other choices can also be used, involving features extracted, not only from high level, but also from mid and lower level layers. From our experiments, such choices  have not proven capable of  significantly improving the achieved performance. 

In the following we present the extraction of concise semantic information, through unsupervised analysis of these representations.

Let us assume that the dataset $S$, including DaTscans and MRI inputs has been collected and used for training the DNN to learn to predict the PD or non-PD status of subjects. Let also  $T$  denote the respective test used to evaluate the performance of the trained network: 
  
\begin{equation}
\label{eq: training dataset}
\mathcal{S} = \big\{(\textbf{x}_{s}(k), y_{s}(k)); \ k=1,\ldots,N_{s}\big\} 
\end{equation}

\begin{equation}
\label{eq: test dataset}
\mathcal{T} = \big\{(\textbf{x}_{t}(k), {y_{t}}(k)); \ k=1,\ldots,N_{t}\big\} 
\end{equation}

In (1), (2), $\textbf{x}_{s}(k)$ and $y_{s}(k)$ denote the $N_{s}$ training inputs and the category to which each one of them belongs. We use a 1 to denote a patient category, and a 0 to denote a control/non-patient one. Similarly, $\textbf{x}_{t}(k)$ and $y_{t}(k)$ denote the $N_{t}$ inputs and the corresponding category over the test set.

Let us assume that we train the DNN using the data in $S$ and, for each input $k$, we collect the $M$ values of the outputs of neurons in the selected DNN fully connected or hidden layer, generating a vector ${\textbf{v}}_{s}(k)$. A similar vector  ${\textbf{v}}_{t}(k)$ is generated when applying the trained DNN to each input $k$:

\begin{equation}
\label{eq:traininglatent}
\mathcal{V}_s = \big\{({\textbf{v}}_{s}(k), \ k=1,\ldots,N_{s}\big\} 
\end{equation}
and 
\begin{equation}
\label{eq:testlatent}
\mathcal{V}_t = \big\{({\textbf{v}}_{t}(k),  \ k=1,\ldots,N_{t}\big\} 
\end{equation}

In the following we derive a concise representation of these $\textbf v$ vectors, by using an unsupervised, clustering procedure.
In particular, we use the k-means++ algorithm  \cite {arthur07} to generate, say, $L$ clusters  ${Q} =\{\textbf{q}_1,\ldots,\textbf{q}_L\}$ through minimisation of the following function: 
\begin{equation}
\label{eq:kmeans}
\widehat{{Q}}_{k\text{-means}} = \underset{{Q}}{\operatorname{arg\ min}} 
\sum_{i=1}^{L} \sum_{\mathbf{v}_{s}\in {V}_{s}}^{} 
\big|\big|\textbf{v}_{s}-\textbf{$\mu$}_{i}\big|\big|^{2}
\end{equation}
in which $\textbf{$\mu$}_{i}$ denotes the mean of $\textbf v$ values belonging to cluster $i$.

For each cluster $i$, we then compute the corresponding cluster center $\textbf{c}(i)$, thus defining the set of cluster centers $C$, which forms a concise representation that can form the proposed prediction model for Parkinson's diagnosis.

\begin{equation}
\label{eq: cluster centroid set}
\mathcal{C} = \big\{(\textbf{c}(i), \ i=1,\ldots,L\big\} 
\end{equation}

This procedure, of using dataset $S$ to generate the set of cluster centers $C$ is illustrated in Fig.3.

\begin{figure}
    \begin{center}
        \includegraphics[scale=0.4]{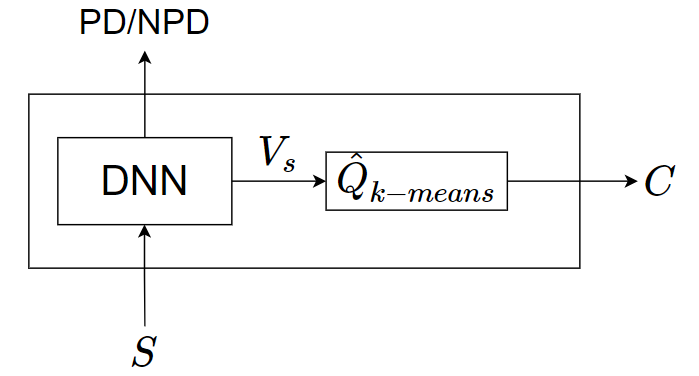}
        \caption{Input set $S$ is used to train the DNN; clustering of the extracted $\mathcal{V}_s$ vector generates  representation set $C$}
        \label{fig:fig3}
    \end{center}
\end{figure}

Since the derived representation consists of a small number of cluster centers, medical experts can examine and annotate the respective DaTscans and MRI images with relevant textual information. This information can include the subject's status (i.e., PD, or non-PD), the stage of Parkinson's for patients, as well as other metrics.

Let us now focus on using the set $C$ for diagnosis of Parkinson's in new subject cases, e.g., those included in the test dataset $T$. For each input in $T$, we compute the 
$\textbf{v}_{s}$ value. We then calculate the euclidean distance of this value from each cluster center in $C$
and classify it to the category of the closest cluster center. As a result, we classify each test input to a respective category, thus predicting the subject's status. 

It should be mentioned that, using this approach, we can predict a new subject's status in a rather efficient and transparent way. At first, only $L$ distances between $M$-dimensional vectors have to be computed and the minimum of them  be selected. Then, the subject can be informed of why the specific diagnosis was made, through visualisation of the medical images and presentation of the medical annotations corresponding to the selected cluster center.

\section{\textbf{The Unified Prediction Model}}

Following the above described approach: a) we design a Deep Neural Network architecture and extensively train it for predicting Parkinson's disease (named DNN in Fig.\ref{fig:fig3}), based on image data provided by a specific hospital, or medical centre, or available database, b) we generate a concise representation (set $C$) composed of the derived cluster center representation that can be used to predict Parkinson's in an efficient way. This
information, i.e., the DNN weights and the set $C$, represent, in the proposed unified approach, the knowledge obtained through the analysis of the respective database $S$.

Let us now consider another medical environment, where another database related to Parkinson's has been generated. Let us assume that it can be, similarly, described through the following training and test sets: 

\begin{equation}
\label{eq: training dataset2}
\mathcal{S'} = \big\{(\textbf{x'}_{s}(k), y'_{s}(k)); \ k=1,\ldots,N'_{s}\big\} 
\end{equation}

\begin{equation}
\label{eq: test dataset2}
\mathcal{T'} = \big\{(\textbf{x'}_{t}(k), {y'_{t}}(k)); \ k=1,\ldots,N'_{t}\big\} 
\end{equation}

In (7), (8),  $\textbf{x'}_{s}(k)$ and $y'_{s}(k)$ denote the $N'_{s}$ training inputs and the corresponding category, whilst $\textbf{x'}_{t}(k)$ and $y'_{t}(k)$ denote the $N'_{t}$
inputs and the corresponding category over the test set.

In the deep learning field it is known that when applying a network, trained on a specific dataset, to another dataset with different characteristics, the performance is expected to be poor. Transfer learning, along with network retraining is the usual technique for obtaining a good performance over the new dataset. However, the 'catastrophic forgetting' problem that was mentioned in the Introduction appears, obstructing the derivation of a unified prediction model over all datasets. 

In the following we show how the proposed approach can alleviate this problem. 

\begin{figure}
    \begin{center}
        \includegraphics[scale=0.40]{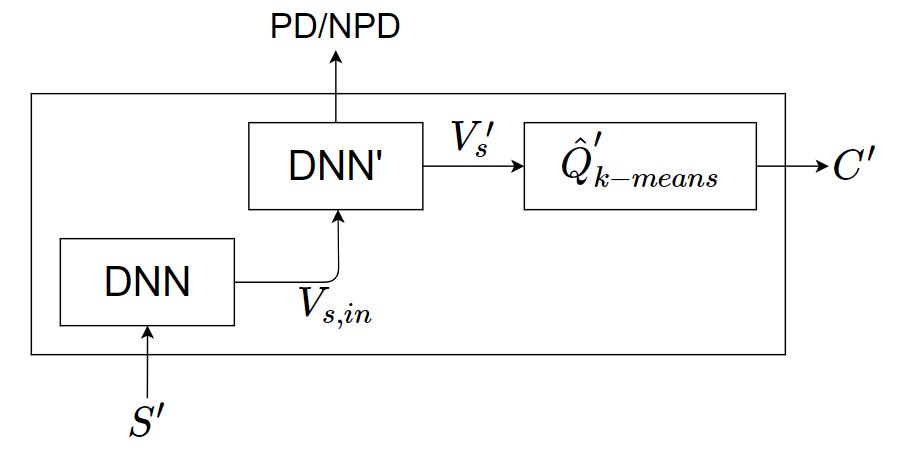}
        \caption{Set $S'$ is fed to DNN, with the extracted ${V}_s$
        vector being used as input for training DNN'; clustering of the extracted ${V'}_s$ vector generates  representation set $C'$}
        \label{fig:fig4}
    \end{center}
\end{figure}

Fig. \ref{fig:fig4} shows the procedure we follow to achieve such a model. According to it, we present all inputs of the new training dataset $S'$ to the available DNN that we have already trained with the original dataset $S$; we compute the $\mathcal {V}_{s}$ representations, similarly to (3), named as $V_{s, in}$ in Fig. \ref{fig:fig4}. These representations, which were generated using the knowledge obtained from the original dataset, form the input to a new DNN, named DNN' in Fig. \ref{fig:fig4}; this network is trained  to use these inputs so as to predict the PD/non-PD status of the subjects whose data are in set $S'$.  

In a similar way, as in (3-5) we compute the new set of representations, named $V'_{s}$ and through clustering the new set of cluster centers $C'$:

\begin{equation}
\label{eq: cluster centroid set2}
\mathcal{C'} = \big\{(\textbf{c'}(i), \ i=1,\ldots,L'\big\} 
\end{equation}

\begin{figure}
    \begin{center}
        \includegraphics[scale=0.40]{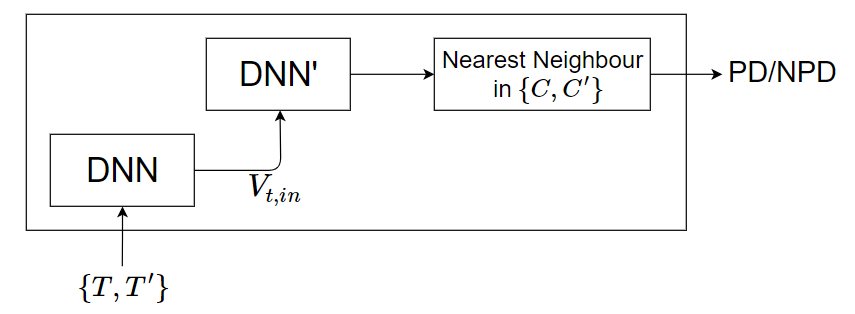}
        \caption{Any subject's data from either set $T$, or $T'$, is fed to the DNN-DNN' architecture, with the extracted $\mathcal{V'}_s$
        vector being classified to the category of the nearest cluster center in $C$ and $C'$; thus predicting subject's status}
        \label{fig:fig5}
    \end{center}
\end{figure}

The next step is to merge the sets $C$ and $C'$, creating the unified prediction model. Using the two network structures (DNN and DNN' in Fig. \ref{fig:fig4}), in a testing formulation, and the nearest neighbor criterion with respect to the union of $C$ and $C'$, we can predict the PD/non-PD status of all subjects in both test sets {T} and {T'}, as  shown in Fig. \ref{fig:fig5}.  

The resulting representation, consisting of the $C$ and $C'$ sets, is, therefore, able to predict a new subject's status, using the knowledge acquired by the DNN and DNN' networks trained on both datasets, in an efficient and transparent way.

\section{\textbf{Domain Adaptation in Parkinson's Prediction}}

In the former Sections it was assumed that the inputs to the deep neural architectures consisted of both DaTscans and MRI data, so that the networks learn to detect and use correlations between both types of inputs. In former research \cite{kollias18,tagaris18} it was shown that DaTscan inputs provide DNNs with more discriminating ability than MRI inputs. Much higher accuracy was achieved by the DNNs when trained with the former than the latter input images.

It should, however, be mentioned that DaTscan facilities are generally available in big medical centers and hospitals. As a consequence, in many medical environments, prediction should be achieved using only MRI information. 

In the following, we present a novel domain adaptation extension of the proposed approach for improving the prediction provided by a DNN when using only MRI inputs, based on the concise $C$ representations derived from a DNN trained with both types of inputs. 

To achieve this, we introduce a novel error criterion for training the new deep neural network with MRI inputs, which is expressed in terms of the internal $\textbf v_{s}$ representations generated by this network, as well as by the representation set $C$ obtained during training of the original network. 

In particular, let us consider that the training and test datasets in (1) and (2) consist of only MRI data. 

By training a DNN with dataset $S$, we can obtain, similarly to (3), a vector $\mathcal{V}''_{s}$, defined as follows:

\begin{equation}
\label{eq:traininglatent2}
\mathcal{V''}_s = \big\{({\textbf{v''}}_{s}(k), \ k=1,\ldots,N''_{s}\big\} 
\end{equation}

where each $\textbf{v''}(s)$ vector is of $M$ dimensionality, equal to the size of the last layer in the CNN or CNN-RNN architecture, for the ${N''_{s}}$ training data. 

Our target is to train the new network to produce $\textbf{v''}_{s}$ values which would be close to one of the cluster centers in $C$ extracted from the original network that was trained with both types of inputs. If this was possible, then the obtained prediction would be closer to the one provided by the original network. As a consequence, a higher prediction accuracy would be obtained by the new network. 

In mathematical terms, we compare, in terms of the mean squared error, the $\textbf{v''}_{s}$
values with the $L$ cluster centers $\textbf{c}_{s}$ defined in (6). 

Based on the minimum euclidean distance criterion, we select a particular cluster center, to form the desired target value for each one of the $\textbf v$'s. As a result, the following $U$ vector of desired values $u(m,n)$is generated:

\begin{equation}
\label{eq:clusterlatent}
{U}_s = \big\{(u(m,n),  m=1,\ldots,L;  n=1,\ldots,N''_{s}\big\} 
\end{equation}

in which $u(m,n)$ equals 1, if the respective cluster center is the selected one among the $L$ dimensional set $C$, or equals 0, if the cluster center is not selected.

The $U_{s}$ values are used in the following to define the new error function, minimisation of which will make the new network tend to duplicate the decision making of the original network and provide improved predictions of the subjects' status.

The proposed error function is composed of two distinct terms. 
The first term is the normal mean squared error criterion computed at the network output level and 
defined as follows:

\begin{equation} \label{eq:5}
\mathcal{F}_1 = \frac{1}{N''_{s}} \sum_{k=1}^{N''_{s}}  (y(k)-z(k))^2 
\end{equation}

in which $z(k)$ represents the category of the input and $y(k)$ represents the respective category prediction provided at the network output.

The following variables are introduced  to define the second term in the error function:

\begin{equation} \label{eq:6}
\textbf{e}(m,n) = \textbf{v''}_{s}(n) - \textbf{c}(m), \  m=1,\ldots,L; \ n=1,\ldots,N''_{s} 
\end{equation}

\begin{equation} \label{eq:7}
E(m,n) = \textbf{e}(m,n)* (\textbf{e}(m,n))^T
\end{equation}
where \textit{T} denotes transposition.

To achieve the targeted goal, we perform minimisation of all $E(m,n)$ values, when $u(m,n)$ equals unity, with simultaneous maximization of the $E(m,n)$ values, when $u(m,n)$ equals zero. Thus, we feed $E(m,n)$ to a nonlinear activation function, of the softmax type, reversing the result, by subtracting it from unity.  

The second error term is computed as the mean squared error between the resulting values and the respective $U_{s}$ ones:

\begin{equation} \label{eq:8}
\mathcal{F}_2 = \frac{1}{LN''_{s}}  \sum_{m=1}^{L} \sum_{n=1}^{N''_{s}} (u(m,n) - [1-f(E(m,n)])^2  
\end{equation}

where \textit{f} is the used softmax function.

Using (12) and (15), the resulting total Error Criterion is computed as follows:

\begin{equation} \label{eq:9}
\mathcal{F}_{new} = \lambda \mathcal{F}_1+ (1-\lambda) \mathcal{F}_2                               
\end{equation}
in which $\lambda$ is a positive number less than unity. When the value of $\lambda$ is close to zero, the significance of the proposed approach is more evident in the obtained results. In general, such a value is used in the following in this paper.

\begin{figure}
    \begin{center}
        \includegraphics[scale=0.30]{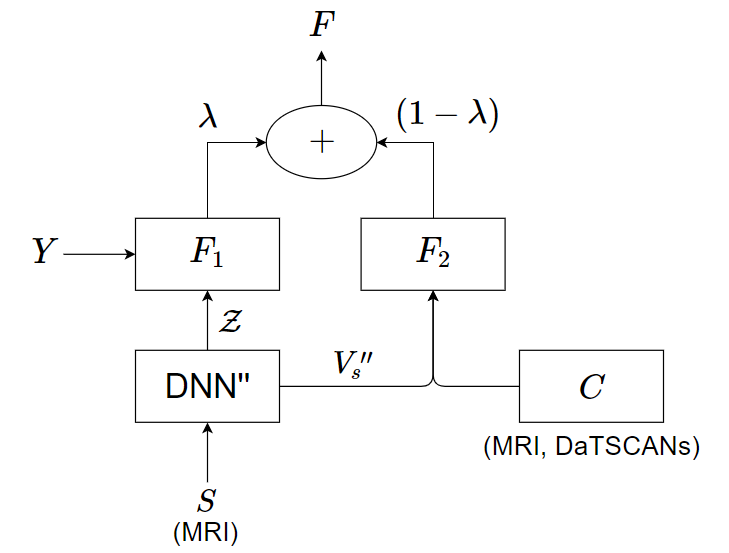}
        \caption{MRI data in $S$ are fed to DNN'' which is trained through minimization of error function $\mathcal{F}$; this is computed based: a) on the difference between output $Z$ and desired output $Y$ ($\mathcal {F}_{1}$ component), b) on comparison of the extracted ${V''}_s$ vector to the representation set {C} computed with both MRI and DaTscan data ($\mathcal {F}_{2}$ component)}
        \label{fig:fig6}
    \end{center}
\end{figure}

Fig. \ref{fig:fig6} presents the proposed training procedure, in which: the set of original cluster centers $C$ is compared to the extracted $V''_{s}$ representations, defining error criterion $F_{2}$;  the DNN'' outputs $Z$, composed of $z(k)$ values, are compared to the desired network predictions $Y$, composed of respective $y(k)$ values, thus defining error criterion $F_{1}$; $F_{1}$ and $F_{2}$ are used to compute the proposed error function minimised during DNN'' network training.

\section{Experimental Study}\label{sec6}

As already described, our experimental study is performed on two databases; the first is the database generated in Greece \cite{tagaris18} and the second is the PPMI database \cite{marek11}. Both of them include DaTscans and MRI information for all their subjects. For training and evaluation purposes the respective datasets have been separated to training, validation and test data. The specific settings can be provided, upon request, from  
\url{mlearn.lincoln.ac.uk}. All experiments have been based on 10-fold cross validation.

Creation of the main deep neural architecture and generation of the respective cluster center representation set for predicting Parkinson's is based on the work  first database. Based on this database we also evaluate the domain adaptation approach for predicting Parkinson's using only MRI information. The unified approach for predicting Parkinson's is based on the data of both databases.

\subsection{Extracting DNN Concise Representations}

In ~\cite{kollias18} deep neural networks were trained with an augmented dataset from database \cite{tagaris18}, achieving very good performance on this database. The convolutional neural network predictor was based on the pretrained ResNet-50 structures, learning to make the prediction in PD/NPD categories through two Fully Connected (FC) layers. It was shown to be able to analyse the spatial characteristics of the DaTscans and MRIs achieving a high accuracy in the database test set, of 94 \%, as shown in Table~\ref{table:results1}. 

\begin{figure*}[ht]
\begin{center}
\includegraphics[scale=0.15]{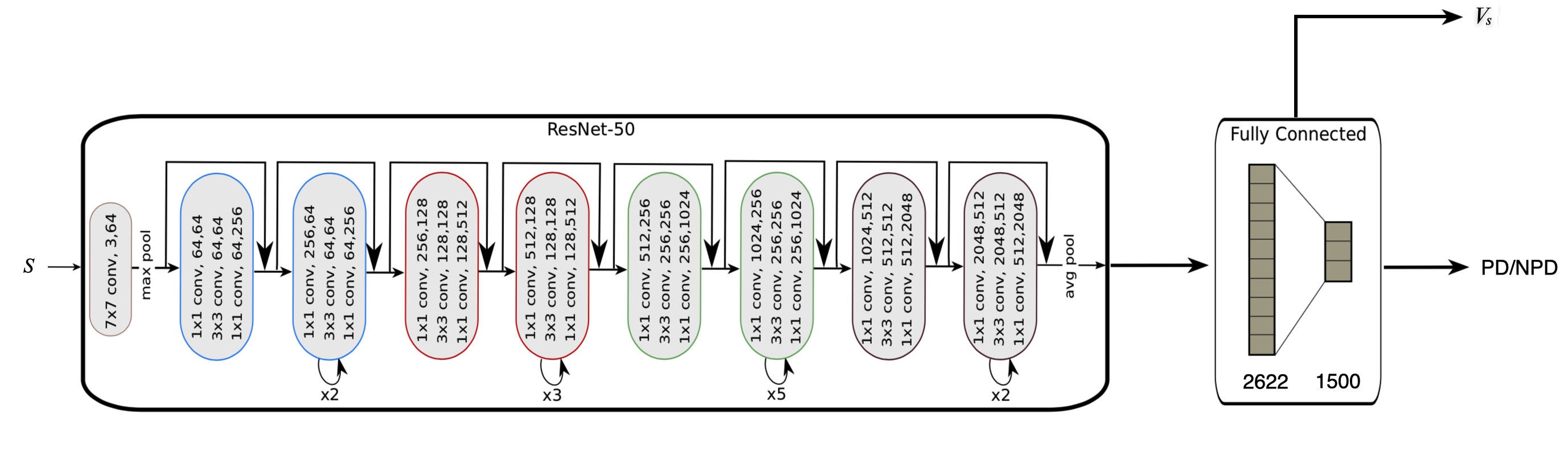}
\end{center}
\caption{A CNN based on the ResNet-50 structure, with two Fully Connected layers; the $\mathcal{V}_{s}$ are extracted from the last FC layer}
\label{fig:resnet_dad}
\end{figure*}

The CNN-RNN architecture included two hidden layers on top of the CNN one, each containing 128 GRU neurons,  as also shown in the Table. This was shown to be able to also analyse the temporal evolution of the MRI data, achieving an improved performance of 98 \% over the test data. We implemented the training using a batch size of 10, a fixed learning rate of 0.001 and a dropout probability of 50 \%.  

\begin{table*}[!t]
\caption{The accuracy obtained by CNN and CNN-RNN architectures}
\label{table:results1}
\centering

\setlength{\arrayrulewidth}{.4pt}
\begin{tabular}{|c|c|c|c|c|c|}
\hline \hline
Structure & No FC layers & No Hidden Layers & No Units in FC Layer(s) & No Units in Hidden Layers & Accuracy ($\%$) \\
\hline
CNN & 2 & - & 2622-1500 & - & {94}\% \\
CNN-RNN & 1 & 2 & 1500 & 128-128 & 98\% \\
\hline \hline
\end{tabular}
\end{table*}

Fig.7 shows the extraction of the CNN $\mathcal{V}_{s}$ vector representations from the last hidden layer of the network, to be further analysed through the clustering procedure. In this case these vectors include 1500 elements, as was shown in  Table~\ref{table:results1}. In the CNN-RNN case, the respective vectors will be extracted from the second RNN hidden layer and will be composed of 128 elements. Since the latter representation has a much lower size than in the CNN case, and at the same time it is able to produce a better prediction accuracy, it is this representation that we use next in this paper.

The clustering process, using the k-means was then applied to the $\mathcal{V}_{s}$ vectors, as shown in Fig.3. 
We extracted five clusters, two of which correspond to control subjects, i.e. NPD ones, with three clusters corresponding to patients, as in the original paper ~\cite{kollias18}. Since the k-means algorithm depends on the initial conditions, the cluster centers are not identical, but very similar to the ones in ~\cite{kollias18}. These constitute the extracted concise representation $C$ set; consequently, $C$ is composed of five 128-dimensional vectors. As was already said, it is the DaTscans that mostly provided the discrimination property to the DNNs. The DaTscans corresponding to the extracted cluster centers are shown in Fig. 8. 

Through the assistance of  medical experts we were able to verify that the three DaTscans corresponding to patient cases represent different stages of Parkinson's disease. In particular: the first of them ($\textbf{c}_{3}$) represents an early occurence, between stage 1 and stage 2; the second ($\textbf{c}_{4}$) shows a pathological case, at stage 2; the third ($\textbf{c}_{5}$) represents a case that has reached stage 3 of Parkinson's. In the case of controls, there are differences between the first ($\textbf{c}_{1}$), which is a clear NPD case and the second ($\textbf{c}_{2}$), which is a more obscure case.

Following the above annotations, it can be said that the derived representations convey more information about the subjects' status than trained DNN outputs. This information can be used by medical experts to evaluate the predictions made by the original DNN when new subjects' data have to be analysed. The computed $\mathcal{V}_{s}$ representations in the new cases can be efficiently classified to the category of the nearest cluster center of $C$; the cluster center's Datscan, MRIs and annotations will then be used to justify, in a transparent way, the provided prediction. 

In Table~\ref{table:results2} we present the amount of training inputs included in every cluster category. Since a large number of cases belong to an early stage of Parkinson's disease, it is of high significance to develop tools, such as the proposed one, which have the ability to provide highly accurate predictions over different datasets and different medical environments.

Let us consider six new subjects, with their data (many combinations of DaTscans and MRIs) having to be analysed by the clustered representation extracted from the trained DNN. There are two NPD and four PD subjects.

We applied the procedure shown in Fig.5 to classify these test data. 
Table~\ref{table:results3} presents the classification of these data to the five generated clusters and consequently to the PD or non-PD category. It can be shown that the proposed approach was able to discriminate all cases, including the early stage Parkinson's cases, with a very high accuracy. This illustrates its ability to provide accurate predictions of Parkinson's disease when provided with DaTscan and MRI data.

Moreover, let us assume that a new case appears, for which; a) the DNN outputs are of low confidence, for example providing output values around 0.5, when a value near to 0 or 1 is required for good prediction; b) the  $\mathcal{V}_{s}$ values are quite faraway from all existing cluster centers in $C$. This means that this is a case that the DNN cannot generalise its learning. As a consequence, a medical expert should annotate these data.

\begin{figure*}[ht]
\begin{center}
\includegraphics[scale=0.5]{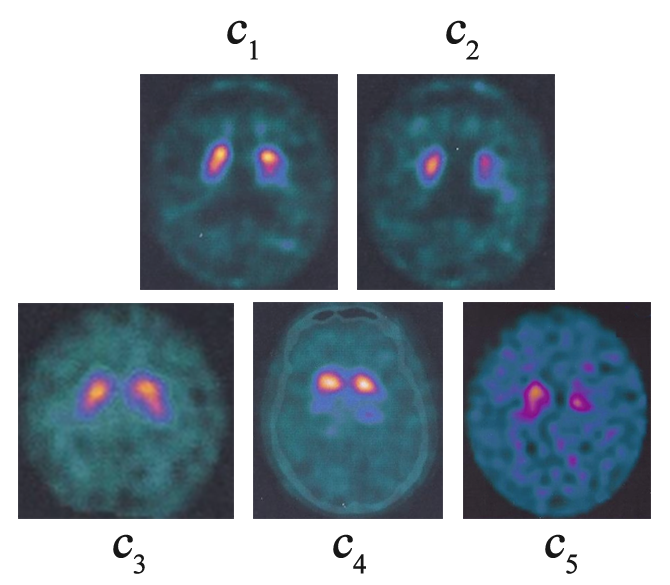}
\end{center}
\caption{The DaTscans of the 5 selected cluster centers: $\textbf{c}_{1}$ and $\textbf{c}_{2}$ correspond to NPD cases, whilst $\textbf{c}_{3}$ - $\textbf{c}_{5}$ to progressing stages of Parkinson's}
\label{fig:scansntua}
\end{figure*}

\begin{table}[!t]
\caption{Percentage of inputs in the five different clusters }
\label{table:results2}
\centering

\setlength{\arrayrulewidth}{.4pt}
\begin{tabular}{|c|c|}
\hline \hline
Cluster & No of Data (\%)  \\
\hline
$\textbf{c}_1$ & 4,3 \\ $\textbf{c}_2$ & 38,4 \\ $\textbf{c}_3$ & 27,6 \\ $\textbf{c}_4$  & 2,3 \\ $\textbf{c}_5$ & 27,4 \\
\hline \hline
\end{tabular}
\end{table}

\begin{table}[!t]
\caption{Test data in each generated cluster and PD/NPD accuracy}
\label{table:results3}
\centering
\setlength{\arrayrulewidth}{.4pt}
\begin{tabular}{|c|c|c|c|c|c|c|}
\hline \hline
Test case & $\textbf{c}_1$ & $\textbf{c}_2$ & $\textbf{c}_3$ & $\textbf{c}_4$ & $\textbf{c}_5$  & PD/NPD \\
\hline
Non Patient 1 & 44 & \textbf{398} & 0 & 0 & 0 & 100 \% \\ 
Non Patient 2 & 10 & \textbf{90} & 0 & 0 & 0 & 100 \% \\
Patient 1 & 3 & 7 & \textbf{94} & 8 & 8 & 91.6 \%\\ 
Patient 2  & 1 & 7 & \textbf{139} & 17 & 20 & 95.6 \% \\ 
Patient 3 & 3 & 0 & \textbf{145} & 18 & 38  & 98.5 \% \\ 
Patient 4 & 0 & 0 & 0 & 8 & \textbf{72} & 100 \%\\
\hline \hline
\end{tabular}
\end{table}

Following the annotation of the new data by the expert, we would need to insert the new data in our prediction system. It should be mentioned that retraining of the deep neural network would  be required, so as to retain the old knowledge and include the new one; this would be computationally intensive and possibly unfeasible. On the contrary, the proposed approach would only require extension of the $C$ set with one, or more, cluster centers, corresponding to the new information; as a consequence, this would be done in a very efficient way.

\subsection{The Unified Prediction Model}

In the following we examine the ability of the proposed approach to generate a unified prediction model for Parkinson's. In particular, we examine the ability of the procedure shown in Fig.4, using the DNN (CNN-RNN) architecture developed in subsection 6.1,  to be successfully applied to the PPMI database \cite{marek11}, for PD/NPD prediction.

Since the DaTscans were the basic source of input information in subsection 6.1, combined with MRI triplets, we focus on the DaTscans included in the PPMI database. For this reason, we have retained 609 subjects from the PPMI database, excluding some patients for which we were not able to extract DaTscans of good quality. In total we selected 1481 DaTscans, which we combined with MRI triplets from the respective subjects, generating a dataset of 7700 inputs; each input was composed, as shown in Fig. 2, of one (gray-scale) DaTscan and a triplet of MRI images.

The data was read in and separated into a training, validation, and testing set each representing about  65 \%, 15 \%, and 20 \% of the set respectively. During separation, care was taken to ensure the split was subject independent. No subject's data were included in more than one set, ensuring that the model learns to solve the problem and not the specific data.
Since the two categories were unbalanced, we performed data augmentation of the NPD category, through addition of small amount of noise, so as to generate a balanced set of 10240 inputs.

At first, for comparison purposes, we trained CNN and CNN-RNN architectures, similar to the ones presented in subsection 6.1, from scratch, on the selected PPMI training set (6656 inputs) and tested the provided accuracy on the test set (2028 inputs), using the validation set (1584 inputs) to test accuracy after completing each training epoch. The obtained accuracy was in the range of 96-97 \%, similar to the accuracy achieved by other techniques, as reported in the Related Work Section of the paper. We also used the pre-trained networks of subsection 6.1, in a transfer learning framework, to initialise the re-training of the new networks. Similar results were obtained in this case as well.

We then applied the procedure described in Section 4 and shown in Fig. 4, to train a DNN' with the $\mathcal{V}_{s}$ vectors extracted from the last hidden layer of the DNN that had been trained on the \cite{tagaris18} dataset. 

We used a  CNN  model, in place of DNN' in Fig. 4. The CNN was fed with the 128-dimensional $\mathcal{V}_{s}$ vectors, and its structure included two Convolution layers, a Max Pooling layer, a Dropout layer with 20 \% probability and three Fully-Connected layers, containing 2688 - 64 -32 neurons respectively, as shown in Fig. 9. As in all DNN training implementations of this paper, we used Python and Tensorflow.

\begin{figure*}[ht]
\begin{center}
\includegraphics[scale=0.6]{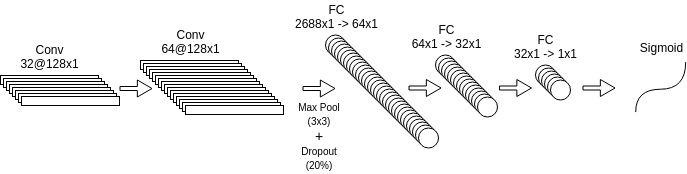}
\end{center}
\caption{A CNN structure, with three Fully Connected layers; the $\mathcal{V}_{s}$ are extracted from the last FC layer }
\label{fig:cnnarchitecture}
\end{figure*}

The performance of the network was very high, classifying in the correct PD/NPD category 99.76 \% of the inputs. The minimization of the Loss function over 500 epochs and the respective accuracy over the test data are shown in Figs. 10 and 11 respectively, while the obtained per class accuracy in all sets, i.e. the training $S'$ and test $T'$ sets, for the PD and NPD categories, is shown in Table 4.

\begin{figure}
    \begin{center}
    \includegraphics[scale=0.50]{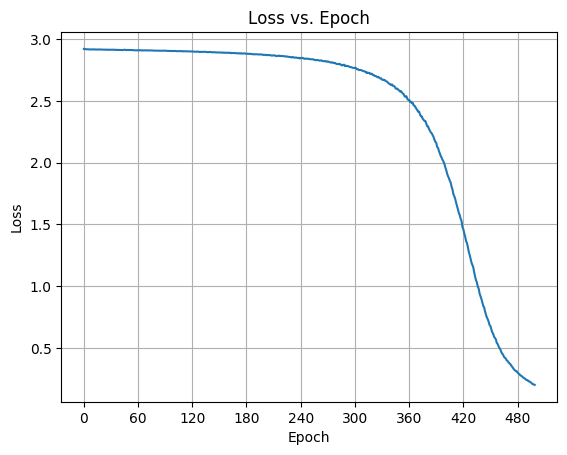}
    \caption{Minimization of the CNN Loss Function in terms of the number of training epochs }
    \label{fig:graphloss}
    \end{center}
\end{figure}

\begin{figure}
    \begin{center}
    \includegraphics[scale=0.50]{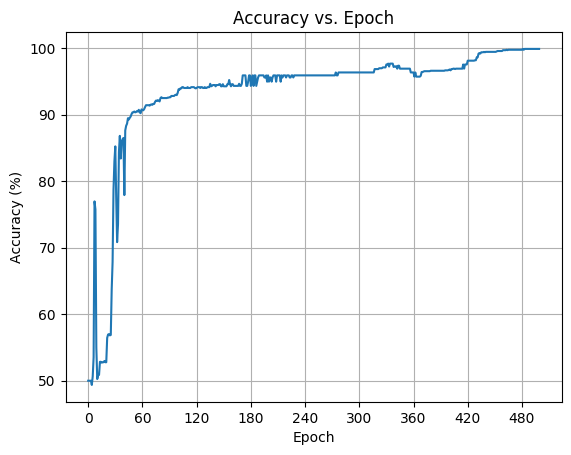}
    \caption{Accuracy (\%) of the CNN when applied to the test dataset in terms of the number of training epochs }
    \label{fig:graphaccu}
    \end{center}
\end{figure}

\begin{table}[!t]
\caption{PD/NPD Accuracy (\%)  on PPMI Dataset}
\label{table:resultsx}
\centering
\begin{tabular}{|c|c|c|c|c|}
\hline \hline
$S'_{PD}$ & $S'_{NPD}$ & $T'_{PD}$ & $T'_{NPD}$  & Total    \\
\hline
99.80 & 99.69 & 99.61  & 99.9 & 99.76 \%\\ 

\hline \hline
\end{tabular}
\setlength{\arrayrulewidth}{.4pt}

\end{table}

By then implementing the clustering procedure shown in Fig.4, we were able to extract five new clusters, three of which represent NPD subjects' cases and two of which represent PD cases. Table 5 presents the split of PPMI data to these five clusters. These cluster centers are 32-dimensional vectors, since they were extracted from the last Fully Connected layer of DNN', which includes 32 neurons.

\begin{table}[!t]
\caption{Percentage of inputs in the new five clusters }
\label{table:resultsy}
\centering

\setlength{\arrayrulewidth}{.4pt}
\begin{tabular}{|c|c|}
\hline \hline
Cluster & No of Data (\%)  \\
\hline
$\textbf{c'}_1$ & 14 \\ $\textbf{c'}_2$ & 13 \\ $\textbf{c'}_3$ & 23 \\ $\textbf{c'}_4$  & 27 \\ $\textbf{c'}_5$ & 23 \\
\hline \hline
\end{tabular}
\end{table}

Fig. 12 shows the DaTscans corresponding to the cluster centers $\textbf{c'}_1$ - $\textbf{c'}_5$. Since the patients in the PPMI Database generally belong to early stages of Parkinson's (stage 1 to stage 2), it can be seen that two cluster centers, i.e., $\textbf{c'}_4$ and $\textbf{c'}_5$ were enough to represent these cases. Variations in the appearance of the non-Parkinson's cases can be seen in $\textbf{c'}_1$ - $\textbf{c'}_3$ DaTscans.

\begin{figure*}[ht]
\begin{center}
\includegraphics[scale=0.4]{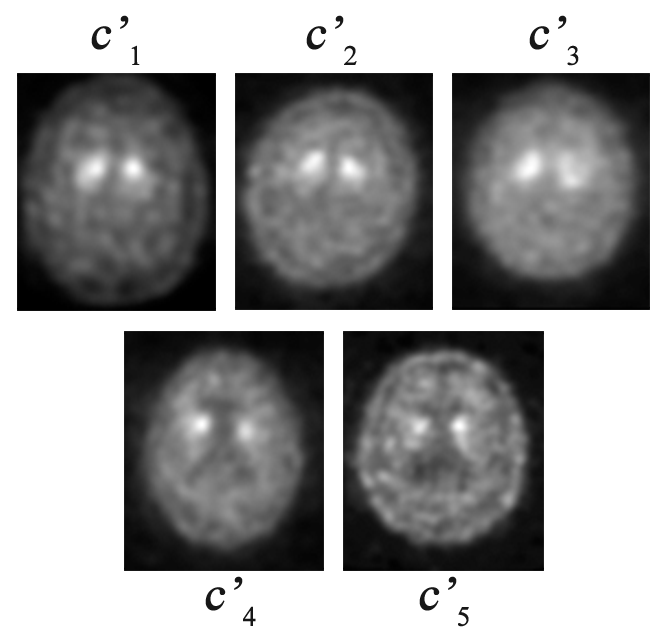}
\end{center}
\caption{The DaTscans of the five cluster centers in $C'$; the three on top represent NPD cases, whilst the two at bottom represent PD cases}
\label{fig:scansppmi}
\end{figure*}

We then applied the merging of sets $C$ and $C'$, as was described in Section 4. 
However, the 5 centers in set $C$ were 128-dimensional, whilst the 5 centers on set $C'$ are 32-dimensional. To produce a unified representation, we made an ablation study, through PCA analysis, on the classification performance achieved in dataset \cite{tagaris18}, if we represented the five cluster centers in $C$ through only 32 principal components. We were able to achieve a classification performance of 97.92 \%, which is very close to the 98 \% performance in Table 1.    

Consequently, we were able to generate a unified model consisting of 10 32-dimensional cluster centers. Fig. 12 shows a 3-D projection of the ten cluster centers. The three (red/rose) squares denote the patient cases in the dataset \cite{tagaris18} and the two (green) plus (+) symbols represent the patient cases in the PPMI dataset. The two (blue) stars represent the normal cases in dataset \cite{tagaris18} and the three (black/grey)  circles represent the normal cases in the PPMI dataset. It can be seen that the PD centers are distinguishable from the NPD ones.

This has been  verified by testing the ability of the unified prediction model to correctly classify all input data in test sets $T$ and $T'$, i.e., the data from both datasets. There was no effect on the performance of the prediction achieved by each prediction model, i.e., $C$ and $C'$ when applied, separately, to their respective datasets, as shown in Tables 1 and 4. This illustrates that the unified representation set, composed of the union of $C$ and $C'$, has been able to provide exactly the same prediction results, as the original representation sets.

\subsection{Domain Adaptation}

If we train the deep neural network described in subsection 6.1 with only MRI data over the dataset \cite{tagaris18}, then the obtained accuracy is just over 70 \%. In particular, if we apply the trained DNN to the six new subject cases of subsection 6.1 and compute the split of the data in the five cluster centers in $C$, the obtained results are shown in Table 6. It can be seen that the prediction, especially in the N-PD cases, is low, with one subject being wrongly classified as PD. 

\begin{figure}
\begin{center}
\includegraphics[scale=0.65]{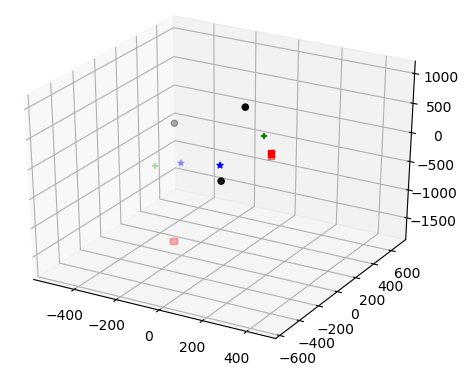}
\end{center}
\caption{The obtained ten cluster centers in 3-D: 5 of them (squares with red/rose color, \&  plus (+) symbols with green color) depict patients; 5 of them (stars with blue color \& circles with black/grey color) depict non-patients}
\label{fig:3dcentroids}
\end{figure}

In the following we examine the application of the domain adaptation approach of Section 5, so as to improve the DNN prediction accuracy when using only MRI inputs.

To do this we implemented the procedure shown in Fig. 6, training the CNN-RNN described in subsection 6.1 with only the MRI training data of the database \cite{tagaris18}. We used the five cluster centers in $C$ to compute the $F_{2}$ error criterion, combining it with the normal mean-squared error criterion $F_{1}$, thus calculating and minimising the total $F$ error criterion.  A value of $\lambda$=0.5 was selected, compensating the contribution of both error components. 

After training, we tested the performance of the adapted DNN over the same test set, obtaining a prediction accuracy of 81.1 \%. Table~\ref{table:results6} illustrates the improvement that was obtained, by using the cluster centers in $C$ as desired values for the extracted $\mathcal{V}_{s}$ values, when compared to the respective results of Table 4. It can be seen that all subjects have been correctly classified to the correct PD/NPD category.

\begin{table}[h]
\caption{Test data in each cluster and PD/NPD accuracy (no adaptation)}
\label{table:results5}
\centering
\begin{tabular}{|c|c|c|c|c|c|c|}
\hline \hline
Test case & $\textbf{c}_1$ & $\textbf{c}_2$ & $\textbf{c}_3$ & $\textbf{c}_4$ & $\textbf{c}_5$ & PD/NPD    \\
\hline
Non Patient 1 & \textbf{181} & 74 & 179 & 8  & 0 & 57.7 \%\\ 
Non Patient 2 & 14 & 4 & \textbf{44}  & 33 & 5 & 25.5 \% \\ 
Patient 1 & 16 & 0  & \textbf{53} & 49 & 2 & 86.7 \% \\ 
Patient 2  & 6 & 0 & \textbf{83} & 80  & 15 & 96.7 \% \\ 
Patient 3 & 26 & 3 & \textbf{130} & 35 & 10 & 85.8 \% \\ 
Patient 4 & 12 & 0 & \textbf{51} & 11 & 6 & 85 \% \\
\hline \hline
\end{tabular}
\setlength{\arrayrulewidth}{.4pt}

\end{table}

\begin{table}[h]
\caption{Test data in each cluster and PD/NPD accuracy (with domain adaptation)}
\label{table:results6}
\centering

\setlength{\arrayrulewidth}{.4pt}
\begin{tabular}{|c|c|c|c|c|c|c|}
\hline \hline
Test case & $\textbf{c}_1$ & $\textbf{c}_2$ & $\textbf{c}_3$ & $\textbf{c}_4$ & $\textbf{c}_5$ & PD/NPD    \\
\hline
Non Patient 1 & \textbf{176} & 147 & 114 & 5 & 0 & 73 \% \\ 
Non Patient 2 & 13 & \textbf{41} & 25 & 18 & 3  & 54 \% \\ 
Patient 1 & 13 & 0 & \textbf{70} & 35  & 2 & 89.2 \%\\ 
Patient 2 & 5 & 0 & \textbf{116} & 54 & 9 & 97.3 \% \\ 
Patient 3 & 20 & 2  & \textbf{140} & 34 & 8 & 89.2 \% \\ 
Patient 4 & 9 & 0 & 31 & 5 & \textbf{35} & 88.8 \% \\
\hline \hline
\end{tabular}
\end{table}

\section{Conclusions and Future Work}

In this paper we have developed a new approach for deriving a unified prediction model for Parkinson's disease.  

We first  extracted concise representations from deep neural networks after training them with DaTscans and MRI data. A set of vectors corresponding to the centers of clusters of these representations, together with the respective DNN structure/weights, constitute the information used to model the knowledge extracted from the PPMI database \cite{marek11} and the Greek database \cite{tagaris18}. 

It has been then shown that the unified model  generated over these different datasets can provide efficient and transparent prediction of Parkinson's disease. Predictions of very high accuracy, which extend the state-of-the-art, have been obtained in both databases.

A domain adaptation methodology, based on the proposed approach was also developed; this introduces a novel error criterion and uses the representations extracted from the DNN that was trained with DaTscans and MRIs, for effectively training respective DNNs in environments that only possess MRI information for their subjects. 

Our future work will follow three directions. 

The first will be to extend the derived unified prediction model for Parkinson's to cover more data cases and be used in real medical environments. We have been collaborating with medical experts and hospitals in Greece and UK for achieving this goal. 

The second will be to extend our former and current research for derivation of a transparent and trustworthy prediction making process; this will include combining the data driven deep neural architectures with knowledge-based methods and ontological representation of knowledge \cite{kollia11}, as well as considering the use of fuzzy descriptors in them \cite{simou08, simou07}. We have been working on extending the early models developed in these works in the current framework of explainable deep learning methodologies.  

The third direction will be to apply the proposed approach to other neurodegenerative diseases, including Alzheimer's disease. Deep learning methodologies have been recently  applied to Alzheimer's data \cite{tagaris18, ortiz16, jo19}. The proposed approach can be applied to these frameworks for unified prediction and for making the deep learning procedure more efficient and transparent.

\section{Acknowledgments}\label{sec11}

{We  thank  Dr Georgios Tagaris and the  Department  of  Neurology
of  the  Georgios  Gennimatas  General  Hospital, Athens, Greece, for providing the dataset with Parkinson's data and for annotating the daTscans corresponding to the extracted cluster center representations. 

The PPMI data  used in the  preparation of   this
 article  were  obtained from  the  Parkinson’s 
Progression   Markers  Initiative  (PPMI)  database
 (www.ppmi-info.org/data). For up-to-date  information
  on   the  study, visit www.ppmi-info.org. PPMI  – a 
public-private partnership –  is funded by the  Michael
J. Fox  Foundation for     Parkinson’s Research and 
funding  partners, including AbbVie, Allergan, Amathus Therapeutics, Avid Radiopharmaceuticals, Biogen, BioLegend, Bristol-Myers Squibb, Celgene, Denali, GE Healthcare, Genentech, GSK, Eli Lilly and Company, Lundbeck, Merck, MSD, Pfizer, Piramal Imaging, Prevail Therapeutics, Roche, Sanofi Genzyme, Servier, .Takeda, Teva, USB, Verily, and Voyager Therapeutics. 

We thank them for providing us with the PPMI dataset used in our experiments to illustrate the performance of the proposed unified prediction model for Parkinson's disease.   

}

\bibliographystyle{iet}
\bibliography{refs.bib}{}

\begin{thebibliography}{10}

\bibitem{goodfellow16}
Goodfellow, I., Bengio, Y., Courville, A.: `Deep Learning'.
\newblock (MIT Press,  2016)

\bibitem{sadja06}
Sadja, P.: `Machine learning for detection and diagnosis of disease',
  \emph{Annual Review of Biomedical Engineering},  2006, pp.~ 537--565

\bibitem{azizi17}
Azizi, S., Bayat, S., Yan, P., Tahmasebi, A.M., Nir, G., Kwak, J.T., et~al.:
  `Detection and grading of prostate cancer using temporal enhanced ultrasound:
  combining deep neural networks and tissue mimicking simulations',
  \emph{International Journal of Computer Assisted Radiology and Surgery},
  2017, pp.~ 1293--1305

\bibitem{li14}
Li, R., Zhang, W., Suk, H., Wang, L., Li, J., Shen, D., et~al.: `Deep learning
  based imaging data completion for improved brain disease diagnosis',
  \emph{International Conference on Medical Image Computing and
  Computer-assisted Intervention},  2014, pp.~ 305--312

\bibitem{goetz08}
Goetz, C.G., Tilley, B.C., Shaftman, S.R., Fahn, S., Martinez.Martin, P.,
  Poewe, W., et~al.: `Movement disorder society-sponsored revision of the
  unified parkinson's disease rating scale (mds-updrs): scale presentation and
  clinimetric testing results', \emph{Movement Disorders},  2008, \textbf{23},
  (15), pp.~2129--2170

\bibitem{hoehn11}
Hoehn, M.M., Yahr, M.D.: `Parkinsonism: Onset, progression, and mortality',
  \emph{Neurology},  1998, \textbf{50}, pp.~318

\bibitem{das10}
Das, R.: `A comparison of multiple classification methods for diagnosis of
  parkinson disease', \emph{Expert Systems with Applications},  2010,
  \textbf{37}, (2), pp.~1568--1572

\bibitem{salvatore14}
Salvatore, C., Cerasa, A., Castiglioni, I., Gallivanone, F., Augimeri, A.,
  Lopez, M., et~al.: `Machine learning on brain mri data for differential
  diagnosis of parkinson's disease and progressive supranuclear palsy',
  \emph{Journal of Neuroscience Methods},  2014, \textbf{222}, pp.~230--237

\bibitem{rojas13}
Rojas, A., Górriz, J.M., Ramírez, J., Illán, I.A., Martínez.Murcia, F.J.,
  Ortiz, A., et~al.: `Application of empirical mode decomposition on datscan
  spect images to explore parkinson disease', \emph{Expert Systems with
  Applications},  2013, \textbf{40}, (7), pp.~2756--2766

\bibitem{kollias18}
Kollias, D., Tagaris, A., Stafylopatis, S., Kollias, S., Tagaris, G.: `Deep
  neural architectures for prediction in healthcare', \emph{Complex \&
  Intelligent Systems},  2018, \textbf{4}, (2), pp.~119--131

\bibitem{tagaris18}
Tagaris, A., Kollias, D., Stafylopatis, A., Tagaris, G., Kollias, S.: `Machine
  learning for neurodegenerative disorder diagnosis - survey of practices and
  launch of benchmark dataset', \emph{International Journal on Artificial
  Intelligent Tools},  2018, \textbf{27}, (3)

\bibitem{tan18}
Tan, C., Sun, F., Kong, T., Zhang, W., Yang, C., Liu, C.: `A survey on deep
  transfer learning', \emph{27th International Conference on Artificial Neural
  Networks},  2018, pp.~ 270--279

\bibitem{kollias18-2}
Kollias, D., Zafeiriou, S.P.: `Training deep neural networks with different
  datasets in-the-wild: The emotion recognition paradigm', \emph{International
  Joint Conference on Neural Networks (IJCNN)},  2018, pp.~ 1--8

\bibitem{marek11}
Marek, K., Jennings, D., Lasch, S., Sideworf, A., Tenner, C., Simuni, T.,
  et~al.: `The {P}arkinson {P}rogression {M}arker {I}nitiative ({PPMI})',
  \emph{Progress in {N}eurobiology},  2011, \textbf{95}, (4), pp.~629--635

\bibitem{prashanth18}
Prashanth, R., Dutta.Roy, S.: `Early detection of parkinson's disease through
  patient questionnaire and predictive modelling', \emph{International Journal
  of Medical Informatics},  2018, \textbf{119}, pp.~75--87

\bibitem{oliveira18}
Oliveira, F.P.M., Faria, D.B., Costa, D.C., Castelo.Branco, M., Tavares,
  J.M.R.S.: `Extraction, selection and comparison of features for an effective
  automated computer-aided diagnosis of parkinson's disease based on
  [123i]fp-cit spect images', \emph{European Journal of Nuclear Medicine and
  Molecular Imaging},  2018, \textbf{45}, (6)

\bibitem{prashanth18-2}
Oliveira, F.P.M., Faria, D.B., Costa, D.C., Castelo.Branco, M., Tavares,
  J.M.R.S.: `Extraction, selection and comparison of features for an effective
  automated computer-aided diagnosis of parkinson's disease based on
  [123i]fp-cit spect images', \emph{International Journal of Medical
  Informatics},  2018, \textbf{119}, pp.~75--87

\bibitem{amoroso18}
Amoroso, N., La.Rocca, M., Monaco, A., Bellotti, R., Tangaro, S.: `Complex
  networks reveal early mri markers of parkinson's disease', \emph{Medical
  Image Analysis},  2018, \textbf{48}, pp.~12--24

\bibitem{singh18}
Singh, G., Samavedham, L., Lim, E.C.: `Determination of imaging biomarkers to
  decipher disease trajectories and differential diagnosis of neurodegenerative
  diseases ({DI}sease {T}re{ND})', \emph{Journal of Neuroscience Methods},
  2018, \textbf{305}, pp.~105--116

\bibitem{lei18}
Lei, H., Huang, Z., Han, T., Luo, Q., Cai, Y., Liu, G., et~al.: `Joint
  regression and classification via relational regularization for parkinson's
  disease diagnosis', \emph{Technology and Healthcare},  2018, \textbf{26},
  pp.~19--30

\bibitem{lei18-2}
Lei, H., Zhao, Y., Wen, Y., Luo, Q., Cai, Y., Liu, G., et~al.: `Sparse feature
  learning for multi-class parkinson's disease classifcation', \emph{Technology
  and Health Care},  2018, \textbf{26}, (S1), pp.~193--203

\bibitem{zhang17}
Zhang, Y., Kagen, A.: `Machine learning interface for medical image analysis',
  \emph{Journal of Digital Imaging},  2017, \textbf{30}, (5), pp.~615--621

\bibitem{tagaris17}
Tagaris, A., Kollias, D., Stafylopatis, A.: `Assessment of parkinson's disease
  based on deep neural networks', \emph{Proceedings of the 17th International
  Conference on Engineering Applications of Neural Networks, Athens, Greece},
  2017, pp.~ 391--403

\bibitem{he16}
He, K., Zhang, X., Ren, S., Sun, J.: `Deep residual learning for image
  recognition', \emph{Proceedings of 2016 IEEE Conference on Computer Vision
  and Pattern Recognition (CVPR)},  2016,

\bibitem{kollias17}
Kollias, D., Yu, M., Tagaris, A., Leontidis, G., Stafylopatis, A., Kollias, S.:
  `Adaptation of contextualization of deep neural network models', \emph{2017
  IEEE Symposium Series on Computational Intelligence (SSCI)},  2017, pp.~ 1--8

\bibitem{ng15}
Ng, H., Nguyen, V.D., Vonikakis, V., Winkler, S.: `Deep learning for emotional
  recognition on small datasets using transfer learning', \emph{Proceedings of
  2015 International Conference on Multimodal Interaction},  2015, pp.~
  443--449

\bibitem{chung14}
Chung, J., Gulcehre, C., Cho, K., Bengio, Y.: `Empirical evaluation of gated
  recurrent neural networks on sequence modeling', \emph{NIPS 2014 Workshop on
  Deep Learning},  2014, pp.~ 1--9

\bibitem{kollia19}
Kollia, I., Stafylopatis, A., Kollias, S.: `Predicting parkinson's disease
  using latent information extracted from deep neural networks', \emph{2019
  IEEE International Joint Conference on Neural Networks (IJCNN)},  2019, pp.~
  1--8

\bibitem{arthur07}
Arthur, D., Vassilvitskii, S.: `K-means++: The advantages of careful seeding',
  \emph{Proceedings of the 18th Annual ACM-SIAM Symposium on Discrete
  Algorithms},  2007, pp.~ 1027--1035

\bibitem{kollia11}
Kollia, I., Glimm, B., Horrocks, I.: `Answering queries over owl ontologies
  with sparql', \emph{OWL ED},  2011,

\bibitem{simou08}
Simou, N., Athanasiadis, T., Stoilos, G., Kollias, S.: `Image indexing and
  retrieval using expressive fuzzy description logics', \emph{Signal, Image and
  Video Processing Journal},  2008, pp.~ 321--335

\bibitem{simou07}
Simou, N., Kollias, S.: `Fi{RE}: A fuzzy reasoning engine for imprecise
  knowledge', ,  2007,

\bibitem{ortiz16}
Ortiz, A., Munilla, J., Górriz, J.M., Ramírez, J.: `Ensembles of deep
  learning architectures for the early diagnosis of the alzheimer's disease',
  \emph{International Journal of Neural Systems},  2016, \textbf{26}, (7)

\bibitem{jo19}
Jo, T., Nho, K., Saykin, A.J.: `Deep learning in alzheimer's disease:
  diagnostic classification and prognostic prediction using neuroimaging data',
  \emph{Frontiers in Aging Neuroscience},  2019, \textbf{11}

\end{thebibliography}

\vfill\pagebreak

\end{document}